\documentclass[conference]{IEEEtran}
\IEEEoverridecommandlockouts
\usepackage{cite}
\usepackage{amsmath,amssymb,amsfonts}
\usepackage{graphicx}
\usepackage{textcomp}
\usepackage{xcolor}
\usepackage{comment}

\def\BibTeX{{\rm B\kern-.05em{\sc i\kern-.025em b}\kern-.08em
    T\kern-.1667em\lower.7ex\hbox{E}\kern-.125emX}}
\begin{document}

\title{Illicit item detection in X-ray images for security applications}

\author{\IEEEauthorblockN{1\textsuperscript{st} Georgios Batsis}
\IEEEauthorblockA{\textit{Department of Informatics and Telematics} \\
\textit{Harokopio University of Athens}\\
Athens, Greece \\
gbatsis@hua.gr}
\and
\IEEEauthorblockN{2\textsuperscript{nd} Ioannis Mademlis}
\IEEEauthorblockA{\textit{Department of Informatics and Telematics} \\
\textit{Harokopio University of Athens}\\
Athens, Greece \\
imademlis@hua.gr}
\and
\IEEEauthorblockN{3\textsuperscript{rd} Georgios Th. Papadopoulos}\hspace{18.7cm}
\IEEEauthorblockA{\textit{Department of Informatics and Telematics} \\
\textit{Harokopio University of Athens}\\
Athens, Greece \\
g.th.papadopoulos@hua.gr}
}

\maketitle

\begin{abstract}
Automated detection of contraband items in X-ray images can significantly increase public safety, by enhancing the productivity and alleviating the mental load of security officers in airports, subways, customs/post offices, etc. The large volume and high throughput of passengers, mailed parcels, etc., during rush hours make it a Big Data analysis task. Modern computer vision algorithms relying on Deep Neural Networks (DNNs) have proven capable of undertaking this task even under resource-constrained and embedded execution scenarios, e.g., as is the case with fast, single-stage, anchor-based object detectors. This paper proposes a two-fold improvement of such algorithms for the X-ray analysis domain, introducing two complementary novelties. Firstly, more efficient anchors are obtained by hierarchical clustering the sizes of the ground-truth training set bounding boxes; thus, the resulting anchors follow a natural hierarchy aligned with the semantic structure of the data. Secondly, the default Non-Maximum Suppression (NMS) algorithm at the end of the object detection pipeline is modified to better handle occluded object detection and to reduce the number of false predictions, by inserting the Efficient Intersection over Union (E-IoU) metric into the Weighted Cluster NMS method. E-IoU provides more discriminative geometrical correlations between the candidate bounding boxes/Regions-of-Interest (RoIs). The proposed method is implemented on a common single-stage object detector (YOLOv5) and its experimental evaluation on a relevant public dataset indicates significant accuracy gains over both the baseline and competing approaches. This highlights the potential of Big Data analysis in enhancing public safety.

\end{abstract}

\begin{IEEEkeywords}
Illicit item detection, X-ray image analysis, Deep Neural Networks, Object detection, Non-Maximum Suppression
\end{IEEEkeywords}

\section{Introduction}
Detecting contraband items using X-ray scanning of luggage, parcels, etc. is a crucial requirement for ensuring public security (e.g. preventing terrorist attacks, fighting smuggling of illegal goods, etc.). X-rays are electromagnetic waves with wavelengths shorter than thee visible light, able to penetrate most materials; X-ray scanners exploit this fundamental property to screen items, such as luggage or packages (e.g., in airports, post/customs offices, etc.). Human operators are able to detect a wide range of potential threats, such as explosives, weapons, or sharp objects, using high-resolution images generated by scanning machines \cite{mery2020x}. However, fully manual screening has important shortcomings: the quality of the scan image can be influenced by several factors, such as occluded objects, cluttered environment or certain material properties of the scanned items \cite{akcay2022towards}, while heavy traffic during rush hours may mentally overload human security officers. Thus, illicit items may be missed, due to the need for ``the line to keep moving" or because of perceptual limitations. The high volume and high throughput of X-ray scans in such scenarios render manual screening ineffective and demand automated Big Data analysis solutions.

Efficient automated X-ray image analysis/screening for automatic illicit item detection is nowadays possible thanks to the advances of computer vision and machine learning. Deep Neural Networks (DNNs) have proven to be remarkably capable in supporting human operators for similar tasks, thus greatly increasing their productivity and reducing the possibility of mistakes. Both whole-image recognition and object detection methods have been proposed for illicit/contraband item detection in X-ray images, based on DNNs. While the former ones simply classify an entire image and assign it an overall class label, algorithms of the latter type identify Regions-of-Interest (RoIs), i.e., bounding boxes that localize (in 2D pixel coordinates) specific objects visible in an input image. While there have been significant advancements in object detection algorithms over the last few decades, achieving sufficient performance in real-world scenarios continues to be a challenge \cite{Thermos_2017_CVPR}. The majority of the proposed methods incorporates mechanisms designed to handle domain-specific aspects (e.g., high occlusions, very cluttered backgrounds, large class imbalance, etc.). Additionally, due to the typically cluttered background of X-ray scan images of luggage, mailed parcels, etc., Non-Maximum Suppression (NMS) is also particularly important for security applications. NMS is the final refinement step incorporated to almost every visual object detection framework, assigned the duty of merging/filtering any spatially overlapping detected RoIs which correspond to a single visible object \cite{Symeonidis2019} \cite{Symeonidis2023}.

Regarding image recognition, the method of \cite{akccay2016transfer} addressed the issue of limited training data by employing a pretrained CNN and fine-tuning it in the X-ray domain, while the method of \cite{miao2019sixray} introduced a module named Class-balanced Hierarchical Refinement (CHR) to enhance the prediction capacity of the CNN under extreme class imbalance. This is an important issue in automated X-ray screening, since negative images (where no illicit item is present) are typically significantly more than the positive ones, with this fact reflected in the relevant available datasets. CHR is separately evaluated on top of three different CNNs: Res-Net101 \cite{he2016deep}, Inception-v3 \cite{szegedy2016rethinking} and Dense-Net \cite{huang2017densely}.

Common DNNs for object detection have also been evaluated with regard to their discrimination capacity and transferability between different X-ray scanners \cite{gaus2019evaluating}; examples include Faster R-CNN \cite{ren2015faster}, Mask R-CNN \cite{he2017mask} and RetinaNet \cite{lin2017focal}. However, modifying fast, anchor-based, single-stage object detectors such as Single Shot MultiBox Detector (SSD) \cite{liu2016ssd} or You Only Look Once (YOLO) \cite{redmon2016you} is the most common approach, due to their ability to operate in real-time even in embedded computer hardware. Such modifications may have various forms. For instance, a Cascaded Structure Tensor (CST) is proposed in \cite{hassan2020cascaded} which took advantage of contour-based information to extract object proposals; the latter ones are then classified using a CNN. An alternative lightweight object detector, called LightRay, is introduced in \cite{ren2022lightray} as a modified version of the YOLOv4 model for small illicit item detection in complex backgrounds. It consisted of a fast MobileNetV3 \cite{howard2019searching} backbone CNN and a feature enhancement network that includes a Lightweight Feature Pyramid Network (LFPN) \cite{lin2017feature}, to obtain information of objects at different scales, and a Convolutional Block Attention Module (CBAM) \cite{woo2018cbam}, for refining feature maps through a spatial attention mechanism.

A different approach was followed in \cite{shao2022exploiting}, where a novel mechanism called Foreground and Background Separation (FBS) was proposed for separating illicit items from complex/cluttered backgrounds. This is achieved by using a feature extraction DNN combined with Spatial Pyramid Pooling (SPP) and a Path Aggregation Network, which extracts high-level features. These feature maps serve as an input to two neural decoders, which reconstruct the background and the foreground simultaneously. Then, an attention module directs the overall model's focus on the foreground objects. In an orthogonal direction, the De-occlusion Attention Module (DOAM) \cite{wei2020occluded} is a neural module designed to overcome occlusion in X-ray images; this is important because occlusions are common, due to the absorption of X-rays by certain materials, such as metals, and the visual overlap of multiple objects within densely packed parcels. DOAM consists of two sub-modules, named Edge Guidance (EG) and Material Awareness (MA), which identify edge and material cues for all visible objects. An alternative domain-specific module is Lateral Inhibition Module (LIM) \cite{tao2021towards}, which includes two components: Bidirectional Propagation (BP) and Boundary Activation (BA). The former one minimizes the impact of neighboring regions, by isolating irrelevant information and the latter one captures object boundaries. Both DOAM and LIM have shown promising results in overcoming object occlusion issues in X-ray scan images.

NMS has also been modified in object detectors for X-ray scan image analysis. For instance, the framework of\cite {zhou2021x} is a modified YOLOv4 detector adopting deformable convolutions \cite{dai2017deformable}, the Gradient Harmonizing Mechanism (GHM) loss \cite{li2019gradient} and an augmented NMS algorithm combining Soft-NMS \cite{bodla2017soft} with the Distance-Intersection-over-Union (DIoU) metric. Focusing on real-time performance, YOLOv5 was modified in \cite{song2022improved} using the Stem \cite{wang2018pelee} and CGhost \cite{han2020ghostnet} modules, resulting in a model with reduced number of parameters that still achieves competitive results in comparison with the baseline method. Finally, the integrated illicit Object Detection (POD) method \cite{ma2023occluded} for X-ray image analysis combines a learnable Gabor layer for edge information retrieval, a spatial attention module for directing focus on low-level features, a Global Context Feature Extraction (GCFE) module and a Dual Scale Feature Aggregation (DSFA) module to enhance semantic information from high-level features.

However, to the best of the authors' knowledge, no object detector devised for the security domain has attempted to modify one basic building block of most single-stage detection frameworks: the anchor boxes. This is particularly important for X-ray screening of luggage or mailed parcels, because better matching between the anchor boxes and the distribution of object sizes/shapes in the training dataset leads to better detection performance on test images. Additionally, despite certain attempts to improve NMS for security applications, the results remain typically sub-optimal under object occlusions, which are common in this domain. Thus, this paper proposes a two-fold improvement of anchor-based, single-stage object detectors for automatically detecting contraband items in X-ray scan images, contributing the following two novelties:
\begin{itemize}
  \item Anchor box optimization by applying Hierarchical Clustering (HC) on the ground-truth object RoIs of the training set. By clustering the ground-truth bounding boxes based on their similarity in terms of size, shape, and position, the resulting clusters can be used to define a natural hierarchy, with larger clusters representing more general object shapes and smaller clusters capturing finer details and variations. The resulting hierarchy can also provide information about the relationships between different object classes.
  \item NMS modification to handle occluded object detection and to reduce false predictions, by computing richer geometrical correlations among candidate RoIs before final bounding box prediction. This is achieved by incorporating the Efficient-IoU metric into the Weighted-Cluster NMS method \cite{zheng2021ciou}.
\end{itemize}

The remainder of the paper is organized as follows. Section \ref{sec:Preliminaries} briefly presents the specific baseline algorithms which is adopted for implementing the proposed novelties (YOLOv5, Weighted-Cluster NMS). Section \ref{sec:method} details the proposed method, consisting of an anchor box refinement approach and a modified NMS algorithm. Section \ref{sec:eval} outlines the experimental evaluation process, which was conducted on a well-known public dataset, and discusses the obtained results. Section \ref{sec:concl} concludes the preceding discussion by identifying the implications of these findings, the limitations of this study and directions for future research.

\section{Preliminaries}
\label{sec:Preliminaries}
In order to evaluate the proposed two-fold method, YOLOv5 \cite{jocher2020yolov5} was adopted as a baseline object detector. The reason behind this choice was solely practical; in principle, the proposed method can be used to augment any other variant of the general anchor-based, single-stage object detection framework, as well.

\subsection{YOLOv5 Architecture}
You Only Look Once (YOLO) \cite{redmon2016you} is a series of fast anchor-based, single-stage object detectors, where object localization and classification are performed using a single CNN. This architecture can, however, be divided into a backbone network, a succeeding neck network and a final prediction head. YOLOv5 \cite{jocher2020yolov5}, which is an update of YOLOv4 \cite{bochkovskiy2020yolov4}, was inspired by EfficientNet \cite{tan2019efficientnet} and, thus, can be easily reconfigured for different network complexity profiles. Out of the common variants (YOLOv5s, YOLOv5m, YOLOv5l, YOLOv5x) the one employed in this paper is YOLOv5l.

\begin{figure}[htbp]
\centerline{\includegraphics[width=\linewidth]{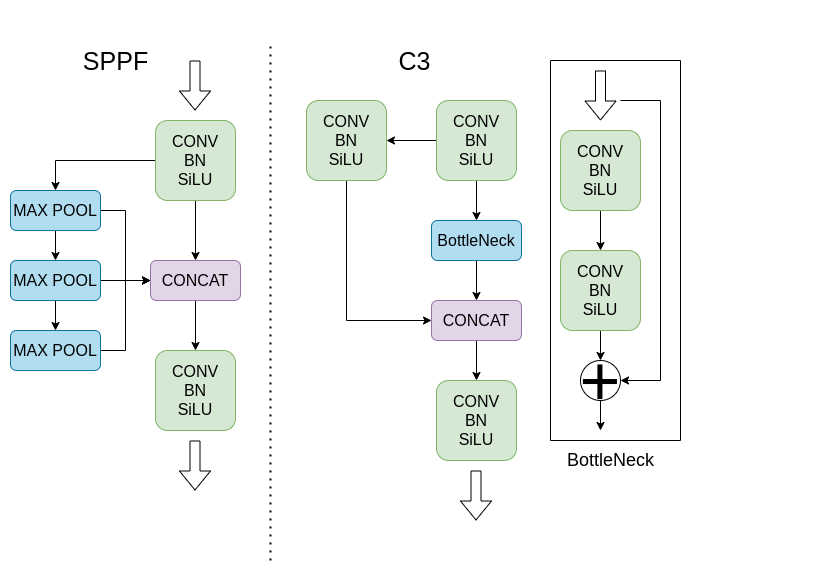}}
\caption{Main YOLO components.}
\label{yolocomp}
\end{figure}

\begin{figure}[htbp]
\centerline{\includegraphics[width=\linewidth]{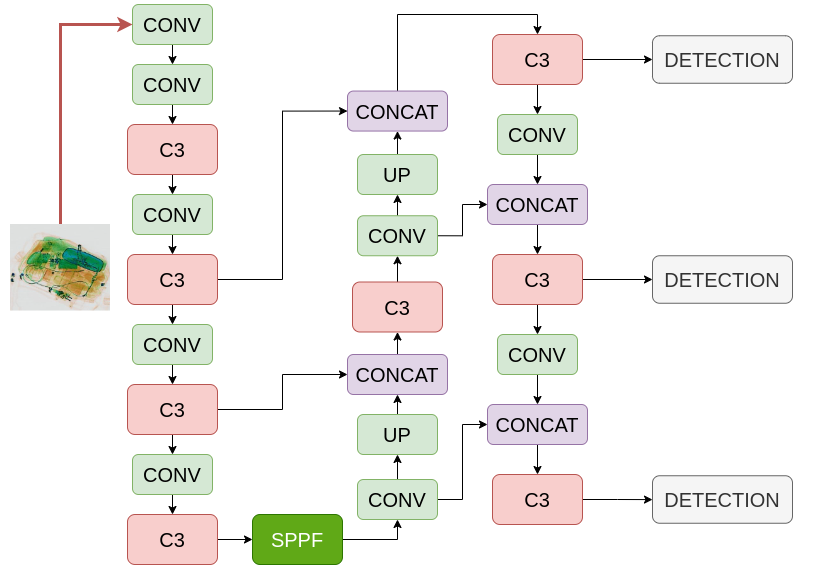}}
\caption{YOLOv5 overall architecture.}
\label{yoloarch}
\end{figure}

The backbone CNN of YOLOv5 is CSP-Darknet53, a modified version of Darknet53 \cite{redmon2018yolov3} combined with Cross Stage Partial (CSP). As presented in Fig. \ref{yolocomp}, the main convolutional block of CSP-Darknet53 consists of convolutional layers, residuals and the SiLU activation function, while the final feature maps are refined using a Spatial Pyramid Pooling-Fast (SPPF) module \cite{he2015spatial}. The neck network consists of a Feature Pyramid Network (FPN) \cite{lin2017feature} and a Path Aggregation Network (PAN) \cite{liu2018path}. These modules repeatedly fuse feature maps from different scales and depth levels, thus leading to final image representations, which are simultaneously characterized by accurate spatial localization details, rich semantics and high invariance regarding object detection. Finally, the prediction head outputs the candidate detected RoIs through a set of convolutional operations. Overall YOLOv5 architecture is presented in Fig. \ref{yoloarch}. 

\subsection{Non-Maximum Suppression}
\begin{figure}[htbp]
\centerline{\includegraphics[width=\linewidth]{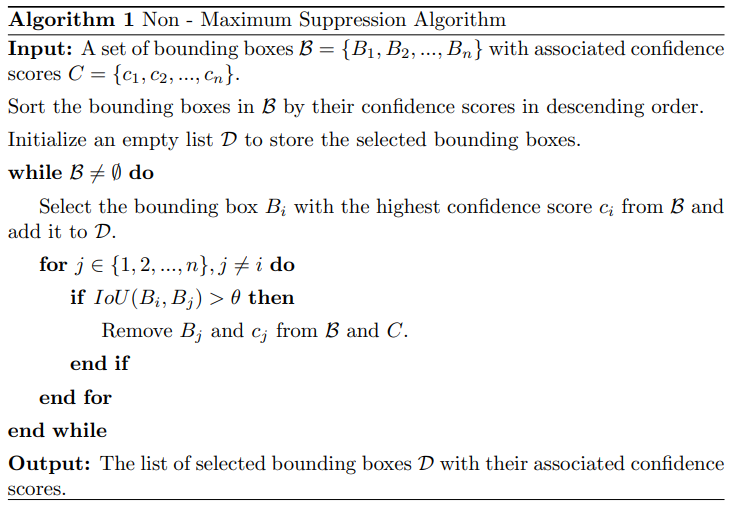}}
\caption{The Greedy Non-Maximum Suppression algorithm.}
\label{nmsAlg}
\end{figure}

Similarly to the majority of object detectors, YOLO generates a large set of overlapping object proposals, in the form of RoIs in pixel coordinates, along with the corresponding class labels and confidence scores. Thus, these candidate RoIs are filtered in a post-processing step based on certain criteria; this is called Non-Maximum Suppression (NMS). The conventional \textit{Greedy NMS} algorithms processes the generated candidate bounding boxes and their corresponding confidence scores for each input image, sorting RoIs in descending confidence order. At first, the box with the highest confidence score is selected and the IoU between itself and all other boxes is calculated. All significantly overlapping RoIs, with an IoU greater than a threshold, are removed. This process is repeated until no bounding boxes remain in the sorted list. The NMS algorithm is presented in Fig. \ref{nmsAlg}.

\section{Proposed Method}
\label{sec:method}
The proposed method is a two-fold improvement of anchor-based, single-stage object detectors, which is highly suitable for the X-ray security scan image analysis domain, due to the peculiarities of such images (e.g., heavy occlusions, cluttered background, etc.).

\subsection{Anchor boxes refinement}
Most anchor-based single-stage object detectors utilize reference anchor boxes of different sizes and aspect ratios, which are placed at various positions across the input image. The goal of these anchor boxes is to capture the variation in object shapes and sizes present in the dataset. Typically, they are predefined (e.g., in the case of YOLOv5, they have been calculated based on prior knowledge of the sizes, aspect ratios, and distributions of ground-truth objects in the COCO dataset \cite{lin2014microsoft}). In many implementations (e.g., YOLOv5) the match between these predefined anchor boxes and the training dataset is verified before training, by computing the achievable recall rate if the object detector using these anchors had access to the ground-truth for all objects in the dataset. If this recall rate is too low, the predefined anchors are assumed to be unfit and a new set of dataset-specific anchor boxes is estimated. This is performed via a run of K-Means to group the ground-truth dataset RoIs into clusters, based on their dimensions in pixel space. The resulting cluster centers are selected as the new anchor boxes, with additional optimizations being possible (e.g., in YOLO a genetic algorithm refines them further). K-Means++ \cite{arthur2006k} can be utilized instead of classic K-Means \cite{luo2022target}.

\begin{figure}[htbp]
\centerline{\includegraphics[width=\linewidth]{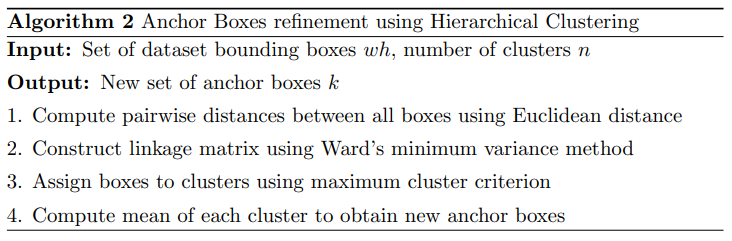}}
\caption{Anchor boxes refinement using Hierarchical Clustering.}
\label{hcAnc}
\end{figure}

In the current work, Hierarchical Clustering (HC) is used to obtain anchor boxes in a dataset-specific manner, according to Algorithm in Fig. \ref{hcAnc}. The goal is to generate anchor boxes that both fit the distribution of ground-truth object sizes/shapes and reflect their arrangement into a natural hierarchy, aligned with the spatial interrelations between the dataset's object classes. Thus, the hierarchy of anchor boxes can provide the detector an explicit template of the dataset's semantic structure, as expressed by the spatial relationships between different object classes. For example, in illicit item detection, RoIs corresponding to classes such as ``knife" and ``wrench" will likely fall under different sub-clusters of a common super-cluster, containing all small-sized handheld items. Since a selection of anchor boxes that better match the dataset's object sizes and shapes is known to lead to better object detection accuracy \cite{luo2022target}, it is reasonable to expect further improvements by obtaining an arrangement of anchors that not only fit the distribution of the dataset's object sizes and shapes, but also reflect the natural hierarchy of the dataset's semantic classes (at least in terms of RoI shape/size).

Algorithm of Fig. \ref{hcAnc} adopts an agglomerative HC method \cite{landau2011cluster} and adapts it to the anchor box refinement task. Its goal is to hierarchically group the training dataset's ground-truth bounding boxes, where each RoI is described as a 2D feature vector: $[w,h]^T$, where $w$/$h$ is the RoI width/height, respectively. First, the pairwise Euclidean distances between all RoIs in the entire training dataset are computed and a linkage matrix is constructed using Ward's minimum variance \cite{ward1963hierarchical}. Then, bounding boxes are assigned to clusters using the maximum cluster criterion and the mean of each cluster is calculated to obtain a new set of corresponding anchor boxes (one per cluster). The total number of target clusters is set to 9, according to experimental evaluation. HC generates a tree-like arrangement of clusters and sub-clusters. The leaves and the root of the clustering tree are not included in the final set of formed anchor boxes.

\subsection{Modified Non-Maximum Suppression}
\begin{figure}[htbp]
\centerline{\includegraphics[width=\linewidth]{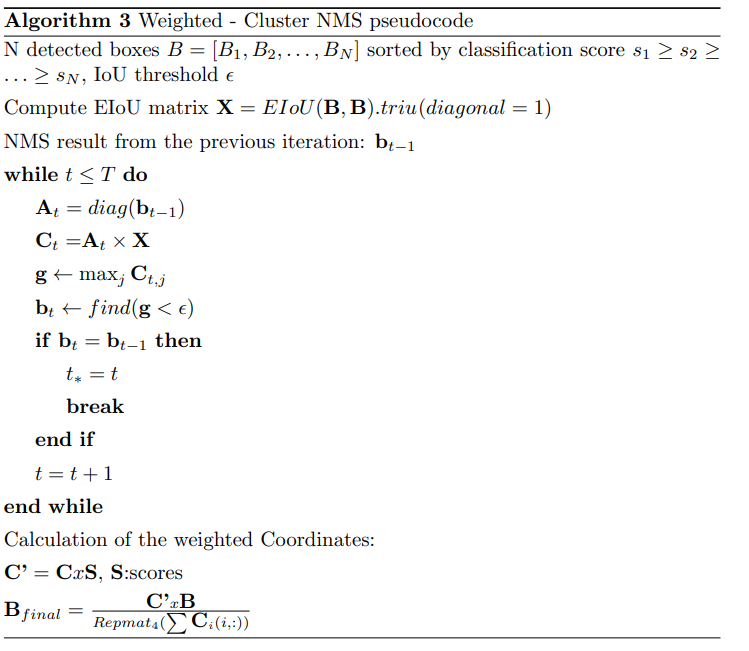}}
\caption{EIoU Weighted-Cluster NMS algorithm.}
\label{wcnms}
\end{figure}

The default Greedy NMS method suffers in the presence of occlusions and gives rise to false positives, triggering various improvements that have been proposed over the years. For instance, Soft-NMS \cite{bodla2017soft} modifies the candidate RoI scores by a Gaussian decay based on the degree of overlap, instead of directly setting the score of all overlapping bounding boxes to zero. This generates more accurate RoIs, even if they are occluded by other objects. Weighted-NMS \cite{zhou2017cad} utilizes the weighted combination of scores and IoU values to define the merged coordinates of the predicted bounding boxes; the result is higher accuracy at the expense of increased time complexity, due to the number of iterations. To mitigate this, Weighted-Cluster NMS (WC-NMS) \cite{zheng2021ciou} has been developed: WC-NMS groups the detected candidate bounding boxes according to the IoU values and then selects the final RoIs according to the maximum score within each group. Implementation-wise this is done with SIMD parallelism and by exploiting cache locality, thanks to formulating the process as a series of matrix operations instead of naive iterative loops, resulting in the fast NMS Algorithm presented in Fig. \ref{wcnms}. Thus, suppression is implemented by calculating the so-called \textit{IoU matrix}. The latter is a symmetric matrix $\textbf{M}$, where $m_{i,j}$ is the IoU between the $i$-th and the $j$-th candidate RoI. Exploiting the symmetry of $\textbf{M}$, Algorithm in Fig. \ref{wcnms} retains only its upper triangular part.

Considering the importance of an efficient NMS method in the X-ray security image analysis domain, due to the densely packed nature of typical luggage and mailed parcels, this paper adopts WC-NMS and improves it, by employing the Efficient-IoU (E-IoU) \cite{zhang2022focal} as the overlap metric. E-IoU is an improvement of Complete-IoU (C-IoU) \cite{zheng2021ciou} and captures richer geometrical information about overlapping candidate bounding boxes, taking into account their overlapping area, their distance and their aspect ratios. Eqs. (\ref{Eq:criterion})-(\ref{Eq:reiou}) define the overlapping criterion according to the so-called \textit{E-IoU matrix}.

\begin{equation}
\mathbf{X} = \mathbf{M}_{IoU} - \mathbf{R}_{EIoU},
\label{Eq:criterion}
\end{equation}

\noindent where $\textbf{M}_{IoU}$ is the IoU matrix, while $\textbf{R}_{EIoU}$ derives from the E-IoU loss \cite{zhang2022focal}. Both matrices are calculated using the predicted candidate RoIs (in top-left, bottom-right pixel coordinates format), as follows:

\begin{equation}
\mathbf{R}_{EIoU} = \frac{\mathbf{D}_{centers}}{(\mathbf{W}^c)^2 + (\mathbf{H}^c)^2} +  \frac{\mathbf{D}^w}{(\mathbf{W}^c)^2} + \frac{\mathbf{D}^h}{(\mathbf{H}^c)^2},
\label{Eq:reiou}
\end{equation}

where , $\textbf{D}_{centers} \in \mathbb{R}^{N \times N}$ is a matrix containing the pairwise Euclidean distances between all $N$ candidate RoIs, with $d_{i,j}$ being the Euclidean distance between the centers of the $i$-th and the $j$-th predicted bounding boxes. $\textbf{W}^c \in \mathbb{R}^{N \times N}$ and $\textbf{H}^c \in \mathbb{R}^{N \times N}$ contain the width and the height, respectively, of the smallest enclosing box covering each pair of candidate ROIs. Thus, $w^{c}_{i,j}$ and $h^{c}_{i,j}$ are the width and height of the smallest bounding box that can contain both the $i$-th and the $j$-th RoI. $\textbf{D}^w \in \mathbb{R}^{N \times N }$ contains the pairwise Euclidean distances between the widths, whereas $\textbf{D}^h \in \mathbb{R}^{N \times N}$ contains the pairwise Euclidean distances between the heights of all $N$ candidate RoIs. Similarly to $\textbf{X}$, all of the matrices mentioned above are symmetric and only their upper triangular part is important for computations and the final result is calculated using element-wise division between the nominator and the denominator, at each term of the sum in Eq. (\ref{Eq:reiou}). Power operations in the denominators are element-wise as well. To sum up, the first term of in Eq. (\ref{Eq:reiou}) captures information about the distance between candidate regions, while the remainder about aspect ratio.

Grouping the candidate RoIs based on their E-IoU values means that rich geometrical information, regarding the spatial interrelations and the arrangement of the detected bounding boxes in 2D pixel space, are inherently considered for grouping them more efficiently. In comparison with D-IoU, which only describes the overlapping area and the distance between candidate boxes, E-IoU captures information about the overlapping area, distance and aspect ratios between the compared RoIs. Although C-IoU also takes into account such geometrical factors, it is computed by estimating the simple difference in aspect ratio between the compared bounding boxes. Such a naive approach may not accurately reflect the actual relationship between the RoI shapes \cite{zhang2022focal}. In order to handle these issues, the second and the third term in Eq. (\ref{Eq:reiou}) capture more sufficiently the similarity between the compared bounding boxes, in terms of their aspect ratio.

\section{Experimental Evaluation}
\label{sec:eval}
This Section overviews the experimental setups used for evaluating the proposed method and discusses the evaluation results. As previously described, YOLOv5 was selected as the baseline to be improved using the proposed method.

\subsection{Experimental Dataset}
The proposed method was implemented and tested using SIXray \cite{miao2019sixray}, a publicly available X-ray security dataset consisting of 1,059,231 X-ray images from subway stations. The 6 classes of illicit objects contained in these images are ``gun", ``knife", ``wrench", ``pliers" and ``scissors". Additionally, a ``negative" class includes all images without any illicit item. Three different dataset subsets are typically utilized in different experimental setups, namely SIXray10, SIXray100 and SIXray1000, where the number indicates the ratio of negative against positive samples. SIXray contains ground-truth whole-image class label annotations manually set by human security inspectors, while their ground-truth object RoIs/bounding boxes are available only for the test set. This paper uses the revised object detection annotations for the training subset provided by \cite{nguyen2022towards}. Despite the fact that only images containing at least one contraband item were utilized, official training-test set split was adopted.


\subsection{Evaluation Metrics}
The effectiveness of the proposed method is measured using the precision, recall and mean Average Precision (mAP) metrics. In object detection tasks, IoU is used to measure the overlap between the predicted and the corresponding ground-truth RoI. In addition, a threshold value was defined in order to decide whether the prediction is actually correct. True Positives (TP), False Positives (FP), and False Negatives (FN) depend on the IoU, the predicted label and the ground-truth label. These elementary metrics are utilized to calculate Precision and Recall:

\begin{equation}
Precision = \frac{TP}{TP+FP}\label{prec}.
\end{equation}

\begin{equation}
Recall = \frac{TP}{TP+FN}\label{rec}.
\end{equation}

\noindent The Precision-Recall (PR) curve depicts the trade-off between precision and recall for different discrimination thresholds. Average Precision (AP) is the area under the PR curve and its range is between 0 to 1. AP is defined as:

\begin{equation}
AP =  \int_{0}^{1} p(r) \,dr\label{AP}.
\end{equation}

\noindent mAP is calculated as the mean of AP over all classes:

\begin{equation}
    mAP = \frac{1}{N} \sum_{i}^{N}AP_i\label{mAP}.
\end{equation}

\subsection{Experimental Evaluation}
Evaluation of all competing methods in the SIXray dataset was conducted using the mAP metric at a 0.5 IoU threshold and the average mAP value at a range of different IoU thresholds. Comparisons were made against the baseline detector implementation before it was augmented with the proposed method (default YOLOv5 with anchor boxes obtained from the COCO dataset and Greedy NMS), as well as with variations using K-Means and K-Means++ clustering for obtaining the anchor boxes, or employing basic (IoU-based) WC-NMS. Additionally, a published YOLOv5 result on SIXray is included for completeness.

Table \ref{blRes} summarizes the accuracy of the baseline method, which achieved a precision of 92.1\%, a recall of 82.3\%, a mAP of 87.6\% and an average mAP across different IoU thresholds (from 0.5 to 0.95) of 72.3\%. Table \ref{methodComp} compares the proposed method against this baseline and against competing approaches based on YOLOv5. The first method is a competing one published in \cite{ma2023occluded}, using YOLOv5 with default anchor boxes, conventional Greedy NMS and a different mini-batch size during training. The next two approaches use Greedy NMS, with one of them employing K-Means-derived and one employing K-Means++-derived anchor boxes. As it can be seen, the proposed method outperforms all other approaches in terms of mAP.

Additionally, Table \ref{ablation} presents an ablation study of the proposed method. The first three variants adopt HC-derived anchor boxes in combination with three different IoU metrics for WC-NMS, namely IoU, D-IoU and C-IoU, respectively. Evidently, D-IoU outperforms the other two metrics in terms of mAP. The last line of Table \ref{methodComp} demonstrates the results of the full proposed method, integrating both HC-based anchors and advanced E-IoU-based WC-NMS into YOLOv5, which outperforms all other (partial) variants in terms of mAP.

\begin{table}[htbp]
\caption{Baseline YOLOv5 results across all classes.}
\begin{center}
\begin{tabular}{|l|l|l|l|l|}
\hline
\textbf{}      & \textbf{Precision} & \textbf{Recall} & \textbf{mAP} & \textbf{mAP50-95} \\ \hline
\textbf{Overall}            & 0.921              & 0.823           & 0.876        & 0.723             \\ \hline
\textbf{Class} &                    &                 & \textbf{AP}  & \textbf{AP50-95}  \\ \hline
Gun            & 0.978              & 0.916           & 0.944        & 0.882             \\ \hline
Knife          & 0.925              & 0.758           & 0.813        & 0.659             \\ \hline
Wrench         & 0.877              & 0.768           & 0.835        & 0.659             \\ \hline
Pliers         & 0.919              & 0.845           & 0.916        & 0.728             \\ \hline
Scissors       & 0.908              & 0.828           & 0.874        & 0.687             \\ \hline
\end{tabular}
\label{blRes}
\end{center}
\end{table}

\begin{table}[htbp]
\caption{Comparative Evaluation.}
\begin{center}
\begin{tabular}{|l|l|l|}
\hline
\textbf{Method}                                           & \textbf{mAP}  & \textbf{mAP50-95} \\ \hline
Baseline                                                  & 87.6          & 72.3              \\ \hline
YOLOv5 baseline of \cite{ma2023occluded} & 86.7          & -                   \\ \hline
K-Means anchors + default NMS                                           & 88            & 73                \\ \hline
K-Means++ anchors + default NMS                                        & 88.3          & 72.9              \\ \hline
\textbf{HC anchors + E-IoU WC-NMS (proposed)}      & \textbf{89.9} & \textbf{75.7}     \\ \hline
\end{tabular}
\label{methodComp}
\end{center}
\end{table}

\begin{table}[htbp]
\caption{Ablation study.}
\begin{center}
\begin{tabular}{|l|l|l|}
\hline
\textbf{Method}                                           & \textbf{mAP}  & \textbf{mAP50-95} \\ \hline
HC anchors + IoU WC-NMS                                 & 89.2          & 75                \\ \hline
HC anchors + D-IoU WC-NMS                                & 89.7          & 75.4              \\ \hline
HC anchors + C-IoU WC-NMS                                & 89.5          & 75.2              \\ \hline
\textbf{HC anchors + E-IoU WC-NMS (full proposed)}      & \textbf{89.9} & \textbf{75.7}     \\ \hline
\end{tabular}
\label{ablation}
\end{center}
\end{table}

\begin{figure}[htbp]
\centerline{\includegraphics[width=\linewidth]{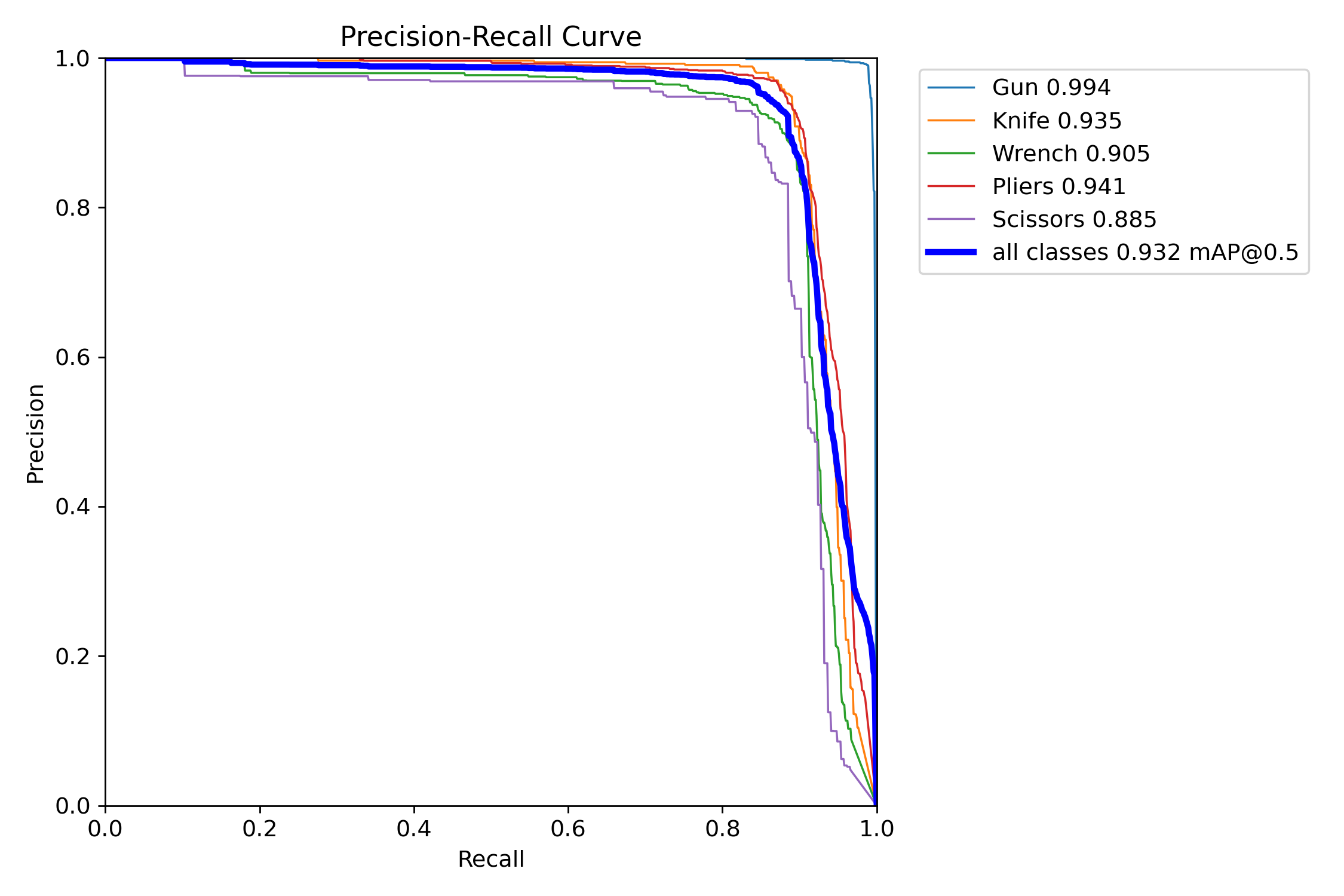}}
\caption{Precision-Recall curve of the proposed method.}
\label{prc}
\end{figure}

\begin{figure}[htbp]
\centerline{\includegraphics[width=1\linewidth]{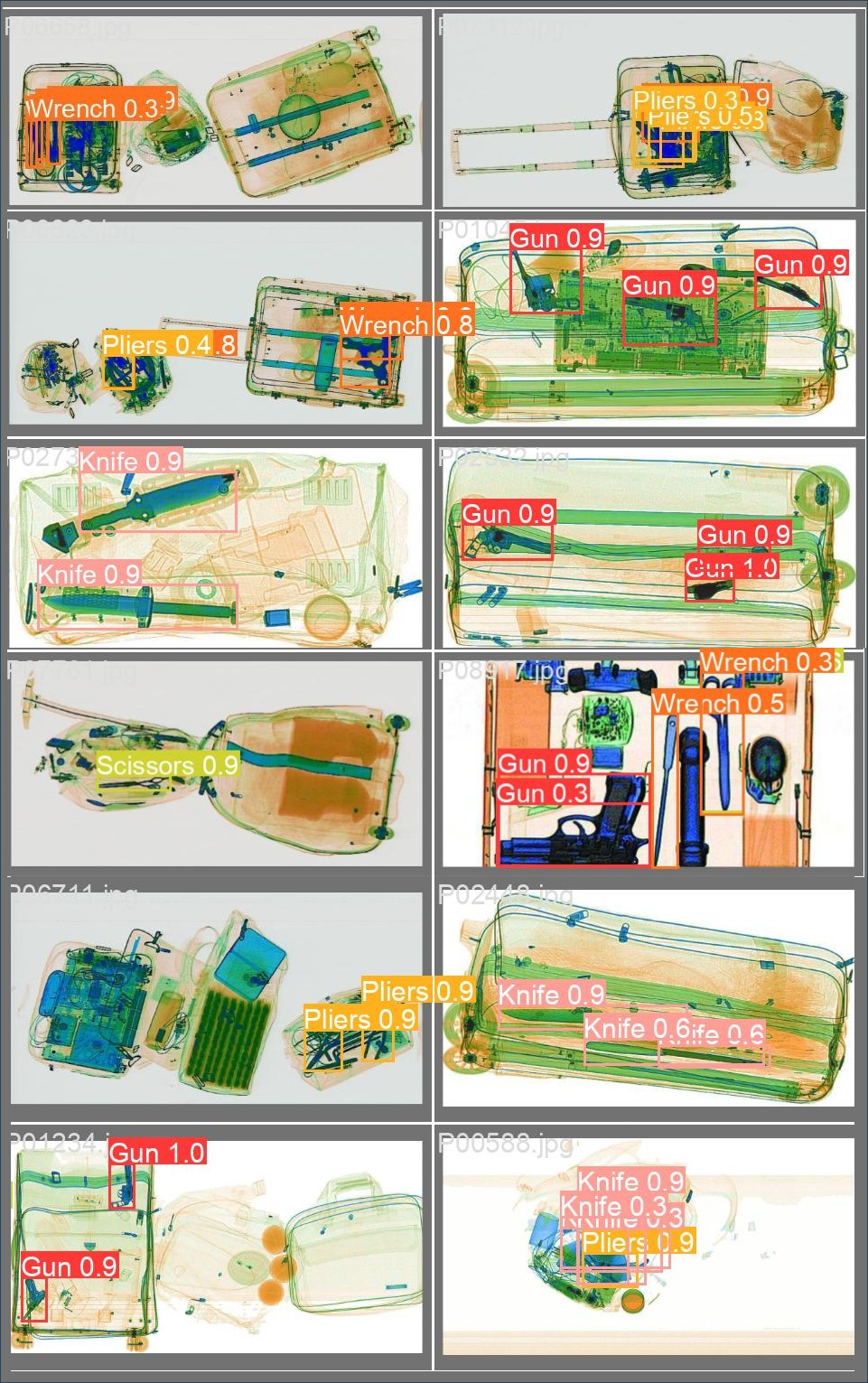}}
\caption{Predictions on the SIXray test subset.}
\label{preds}
\end{figure}

Table \ref{finalRes} presents the complete evaluation of the proposed method across all classes, using HC-derived anchor boxes and the E-IoU-based WC-NMS. It outperforms the default YOLOv5-large by 2.3\% in terms of mAP. Its mAP is 89.9\% at a 0.5 IoU threshold, while the average mAP across a range of different IoU thresholds is 75.7\%. The proposed method significantly reduces the number of false predictions and is more accurate in detecting contraband items, especially in cases of occluded object detection. In Figure \ref{prc}, the precision-recall curve of the proposed framework is presented, which highlights the performance of the model in the desired task. The curve shows a high precision score for low recall values, indicating that the model is very selective in its predictions. However, as recall increases, the precision score decreases, suggesting that the model struggles with correctly classifying some samples. However, the Area Under the Curve (AUC) value indicates that the model performs robust predictions. The above findings suggest that the model exhibits potentials for use in specific applications where high precision is critical, such as contraband detection. Finally, our model was deployed in the test subset of SIXray dataset and the predictions are presented in Fig. \ref{preds}. Notably, neither HC-derived anchor boxes nor E-IoU-based WC-NMS have been proposed/investigated before for object detection.

\begin{table}[htbp]
\caption{Accuracy of the proposed method across all classes.}
\begin{center}
\begin{tabular}{|l|l|l|l|l|}
\hline
\textbf{Class} & \textbf{Precision} & \textbf{Recall} & \textbf{mAP} & \textbf{mAP 50-95} \\ \hline
Overall        & 0.949              & 0.837           & 0.899        & 0.757              \\ \hline
               &                    &                 & \textbf{AP}  & \textbf{AP 50-95}  \\ \hline
Gun            & 0.977              & 0.945           & 0.971        & 0.917              \\ \hline
Knife          & 0.954              & 0.789           & 0.841        & 0.692              \\ \hline
Wrench         & 0.904              & 0.781           & 0.86         & 0.695              \\ \hline
Pliers         & 0.955              & 0.833           & 0.924        & 0.75               \\ \hline
Scissors       & 0.956              & 0.84            & 0.899        & 0.73               \\ \hline
\end{tabular}
\label{finalRes}
\end{center}
\end{table}

\section{Conclusions}
\label{sec:concl}
The large volume and high throughput of passengers or mailed parcels during rush hours, in airports, subways or post/customs offices, make the automated detection of contraband items in X-ray images a Big Data analysis task that is critical for public safety. This paper proposed a novel approach to improve the performance of single-stage, anchor-based object detectors in the X-ray domain. It incorporated two complementary improvements: dataset-specific hierarchical clustering of ground-truth training RoIs, so that the derived anchor boxes better match the distribution and semantic hierarchy of object sizes/shapes, and a modification of an efficient NMS algorithm, so as to better handle occluded objects and to reduce false predictions. According to a thorough experimental evaluation on a relevant public dataset, the proposed method outperforms both the baseline and various competing approaches.

Future research will focus on addressing other limitations of existing methods, such as low generalization ability under domain shifts (e.g., if the detector has been trained with X-rays from one type of scanner and is deployed in an airport using a different scanner), as well as integration of the proposed approach to more recent object detectors (e.g., YOLOv7).

\section*{Acknowledgment}
The research leading to these results has received funding from the European Union's Horizon Europe research and innovation programme under grant agreement No 101073876 (Ceasefire). This publication reflects only the authors views. The European Union is not liable for any use that may be made of the information contained therein.

\bibliographystyle{IEEEtran}
\bibliography{bibliography.bib}

\begin{thebibliography}{10}
\providecommand{\url}[1]{#1}
\csname url@samestyle\endcsname
\providecommand{\newblock}{\relax}
\providecommand{\bibinfo}[2]{#2}
\providecommand{\BIBentrySTDinterwordspacing}{\spaceskip=0pt\relax}
\providecommand{\BIBentryALTinterwordstretchfactor}{4}
\providecommand{\BIBentryALTinterwordspacing}{\spaceskip=\fontdimen2\font plus
\BIBentryALTinterwordstretchfactor\fontdimen3\font minus
  \fontdimen4\font\relax}
\providecommand{\BIBforeignlanguage}[2]{{%
\expandafter\ifx\csname l@#1\endcsname\relax
\typeout{** WARNING: IEEEtran.bst: No hyphenation pattern has been}%
\typeout{** loaded for the language `#1'. Using the pattern for}%
\typeout{** the default language instead.}%
\else
\language=\csname l@#1\endcsname
\fi
#2}}
\providecommand{\BIBdecl}{\relax}
\BIBdecl

\bibitem{mery2020x}
D.~Mery, D.~Saavedra, and M.~Prasad, ``{X}-ray baggage inspection with computer
  vision: A survey,'' \emph{IEEE Access}, vol.~8, pp. 145\,620--145\,633, 2020.

\bibitem{akcay2022towards}
S.~Akcay and T.~Breckon, ``Towards automatic threat detection: A survey of
  advances of deep learning within {X}-ray security imaging,'' \emph{Pattern
  Recognition}, vol. 122, p. 108245, 2022.

\bibitem{Thermos_2017_CVPR}
S.~Thermos, G.~T. Papadopoulos, P.~Daras, and G.~Potamianos, ``Deep
  affordance-grounded sensorimotor object recognition,'' in \emph{Proceedings
  of the IEEE Conference on Computer Vision and Pattern Recognition (CVPR)},
  July 2017.

\bibitem{Symeonidis2019}
C.~Symeonidis, I.~Mademlis, N.~Nikolaidis, and I.~Pitas, ``Improving neural
  {Non-Maximum Suppression} for object detection by exploiting interest-point
  detectors,'' in \emph{Proceedings of the IEEE International Workshop on
  Machine Learning for Signal Processing (MLSP)}, 2019.

\bibitem{Symeonidis2023}
C.~Symeonidis, I.~Mademlis, I.~Pitas, and N.~Nikolaidis, ``Neural
  attention-driven {Non-Maximum Suppression} for person detection,'' \emph{IEEE
  Transactions on Image Processing}, 2023, accepted for publication.

\bibitem{akccay2016transfer}
S.~Ak{\c{c}}ay, M.~E. Kundegorski, M.~Devereux, and T.~P. Breckon, ``Transfer
  learning using convolutional neural networks for object classification within
  {X}-ray baggage security imagery,'' in \emph{Proceedings of the IEEE
  International Conference on Image Processing (ICIP)}, 2016.

\bibitem{miao2019sixray}
C.~Miao, L.~Xie, F.~Wan, C.~Su, H.~Liu, J.~Jiao, and Q.~Ye, ``{SIXray}: A
  large-scale security inspection {X}-ray benchmark for prohibited item
  discovery in overlapping images,'' in \emph{Proceedings of the IEEE/CVF
  Conference on Computer Vision and Pattern Recognition (CVPR)}, 2019.

\bibitem{he2016deep}
K.~He, X.~Zhang, S.~Ren, and J.~Sun, ``Deep residual learning for image
  recognition,'' in \emph{Proceedings of the IEEE Conference on Computer Vision
  and Pattern Recognition (CVPR)}, 2016.

\bibitem{szegedy2016rethinking}
C.~Szegedy, V.~Vanhoucke, S.~Ioffe, J.~Shlens, and Z.~Wojna, ``Rethinking the
  inception architecture for computer vision,'' in \emph{Proceedings of the
  IEEE Conference on Computer Vision and Pattern Recognition (CVPR)}, 2016.

\bibitem{huang2017densely}
G.~Huang, Z.~Liu, L.~Van Der~Maaten, and K.~Q. Weinberger, ``Densely connected
  convolutional networks,'' in \emph{Proceedings of the IEEE Conference on
  Computer Vision and Pattern Recognition (CVPR)}, 2017.

\bibitem{gaus2019evaluating}
Y.~F.~A. Gaus, N.~Bhowmik, S.~Akcay, and T.~Breckon, ``Evaluating the
  transferability and adversarial discrimination of convolutional neural
  networks for threat object detection and classification within {X}-ray
  security imagery,'' in \emph{Proceedings of the IEEE International Conference
  On Machine Learning And Applications (ICMLA)}, 2019.

\bibitem{ren2015faster}
S.~Ren, K.~He, R.~Girshick, and J.~Sun, ``Faster {R-CNN}: Towards real-time
  object detection with region proposal networks,'' \emph{Proceedings of the
  Advances in Neural Information Processing Systems (NIPS)}, 2015.

\bibitem{he2017mask}
K.~He, G.~Gkioxari, P.~Dollár, and R.~Girshick, ``Mask r-cnn,'' in \emph{2017
  IEEE International Conference on Computer Vision (ICCV)}, 2017, pp.
  2980--2988.

\bibitem{lin2017focal}
T.-Y. Lin, P.~Goyal, R.~Girshick, K.~He, and P.~Doll{\'a}r, ``Focal loss for
  dense object detection,'' in \emph{Proceedings of the IEEE International
  Conference on Computer Vision (ICCV)}, 2017.

\bibitem{liu2016ssd}
W.~Liu, D.~Anguelov, D.~Erhan, C.~Szegedy, S.~Reed, C.-Y. Fu, and A.~C. Berg,
  ``Ssd: Single shot multibox detector,'' in \emph{Computer Vision--ECCV 2016:
  14th European Conference, Amsterdam, The Netherlands, October 11--14, 2016,
  Proceedings, Part I 14}.\hskip 1em plus 0.5em minus 0.4em\relax Springer,
  2016, pp. 21--37.

\bibitem{redmon2016you}
J.~Redmon, S.~Divvala, R.~Girshick, and A.~Farhadi, ``You only look once:
  Unified, real-time object detection,'' in \emph{Proceedings of the IEEE
  Conference on Computer Vision and Pattern Recognition (CVPR)}, 2016.

\bibitem{hassan2020cascaded}
T.~Hassan, S.~Akcay, M.~Bennamoun, S.~Khan, and N.~Werghi, ``Cascaded structure
  tensor framework for robust identification of heavily occluded baggage items
  from {X}-ray scans,'' \emph{arXiv preprint arXiv:2004.06780}, 2020.

\bibitem{ren2022lightray}
Y.~Ren, H.~Zhang, H.~Sun, G.~Ma, J.~Ren, and J.~Yang, ``{LightRay}: Lightweight
  network for prohibited items detection in {X}-ray images during security
  inspection,'' \emph{Computers and Electrical Engineering}, vol. 103, p.
  108283, 2022.

\bibitem{howard2019searching}
A.~Howard, M.~Sandler, G.~Chu, L.-C. Chen, B.~Chen, M.~Tan, W.~Wang, Y.~Zhu,
  R.~Pang, V.~Vasudevan \emph{et~al.}, ``Searching for {MobileNetv3},'' in
  \emph{Proceedings of the IEEE/CVF International Conference on Computer Vision
  (ICCV)}, 2019.

\bibitem{lin2017feature}
T.-Y. Lin, P.~Doll{\'a}r, R.~Girshick, K.~He, B.~Hariharan, and S.~Belongie,
  ``Feature pyramid networks for object detection,'' in \emph{Proceedings of
  the IEEE conference on computer vision and pattern recognition}, 2017, pp.
  2117--2125.

\bibitem{woo2018cbam}
S.~Woo, J.~Park, J.-Y. Lee, and I.~S. Kweon, ``Cbam: Convolutional block
  attention module,'' in \emph{Proceedings of the European conference on
  computer vision (ECCV)}, 2018, pp. 3--19.

\bibitem{shao2022exploiting}
F.~Shao, J.~Liu, P.~Wu, Z.~Yang, and Z.~Wu, ``Exploiting foreground and
  background separation for prohibited item detection in overlapping {X-ray}
  images,'' \emph{Pattern Recognition}, vol. 122, p. 108261, 2022.

\bibitem{wei2020occluded}
Y.~Wei, R.~Tao, Z.~Wu, Y.~Ma, L.~Zhang, and X.~Liu, ``Occluded prohibited items
  detection: An {X}-ray security inspection benchmark and de-occlusion
  attention module,'' in \emph{Proceedings of the ACM International Conference
  on Multimedia (ACM MM)}, 2020.

\bibitem{tao2021towards}
R.~Tao, Y.~Wei, X.~Jiang, H.~Li, H.~Qin, J.~Wang, Y.~Ma, L.~Zhang, and X.~Liu,
  ``Towards real-world {X}-ray security inspection: A high-quality benchmark
  and lateral inhibition module for prohibited items detection,'' in
  \emph{Proceedings of the IEEE/CVF International Conference on Computer Vision
  (ICCV)}, 2021.

\bibitem{zhou2021x}
C.~Zhou, H.~Xu, B.~Yi, W.~Yu, and C.~Zhao, ``X-ray security inspection image
  detection algorithm based on improved {YOLOv4},'' in \emph{Proceedings of the
  IEEE Eurasia Conference on IOT, Communication and Engineering (ECICE)}, 2021.

\bibitem{dai2017deformable}
J.~Dai, H.~Qi, Y.~Xiong, Y.~Li, G.~Zhang, H.~Hu, and Y.~Wei, ``Deformable
  convolutional networks,'' in \emph{Proceedings of the IEEE International
  Conference on Computer Vision (ICCV)}, 2017.

\bibitem{li2019gradient}
B.~Li, Y.~Liu, and X.~Wang, ``Gradient harmonized single-stage detector,'' in
  \emph{Proceedings of the AAAI Conference on Artificial Intelligence (AAAI)},
  2019.

\bibitem{bodla2017soft}
N.~Bodla, B.~Singh, R.~Chellappa, and L.~S. Davis, ``Soft-{NMS}: improving
  object detection with one line of code,'' in \emph{Proceedings of the IEEE
  International Conference on Computer Vision (ICCV)}, 2017.

\bibitem{song2022improved}
B.~Song, R.~Li, X.~Pan, X.~Liu, and Y.~Xu, ``Improved {YOLOv5} detection
  algorithm of contraband in x-ray security inspection image,'' in
  \emph{Proceedings of the International Conference on Pattern Recognition and
  Artificial Intelligence (PRAI)}.\hskip 1em plus 0.5em minus 0.4em\relax IEEE,
  2022.

\bibitem{wang2018pelee}
R.~J. Wang, X.~Li, and C.~X. Ling, ``Pelee: A real-time object detection system
  on mobile devices,'' \emph{Proceedings of Advances in Neural Information
  Processing Systems (NIPS)}, 2018.

\bibitem{han2020ghostnet}
K.~Han, Y.~Wang, Q.~Tian, J.~Guo, C.~Xu, and C.~Xu, ``Ghostnet: More features
  from cheap operations,'' in \emph{Proceedings of the IEEE/CVF Conference on
  Computer Vision and Pattern Recognition (CVPR)}, 2020.

\bibitem{ma2023occluded}
C.~Ma, L.~Zhuo, J.~Li, Y.~Zhang, and J.~Zhang, ``Occluded prohibited object
  detection in {X}-ray images with global context-aware multi-scale feature
  aggregation,'' \emph{Neurocomputing}, vol. 519, pp. 1--16, 2023.

\bibitem{zheng2021ciou}
Z.~Zheng, P.~Wang, D.~Ren, W.~Liu, R.~Ye, Q.~Hu, and W.~Zuo, ``Enhancing
  geometric factors in model learning and inference for object detection and
  instance segmentation,'' 2021.

\bibitem{jocher2020yolov5}
\BIBentryALTinterwordspacing
G.~Jocher, ``{YOLOv5 by Ultralytics}.'' [Online]. Available:
  \url{https://github.com/ultralytics/yolov5}
\BIBentrySTDinterwordspacing

\bibitem{bochkovskiy2020yolov4}
A.~Bochkovskiy, C.-Y. Wang, and H.-Y.~M. Liao, ``{YOLO}v4: Optimal speed and
  accuracy of object detection,'' \emph{arXiv preprint arXiv:2004.10934}, 2020.

\bibitem{tan2019efficientnet}
M.~Tan and Q.~Le, ``Efficientnet: Rethinking model scaling for convolutional
  neural networks,'' in \emph{International conference on machine
  learning}.\hskip 1em plus 0.5em minus 0.4em\relax PMLR, 2019, pp. 6105--6114.

\bibitem{redmon2018yolov3}
J.~Redmon and A.~Farhadi, ``{YOLO}v3: An incremental improvement,'' \emph{arXiv
  preprint arXiv:1804.02767}, 2018.

\bibitem{he2015spatial}
K.~He, X.~Zhang, S.~Ren, and J.~Sun, ``Spatial pyramid pooling in deep
  convolutional networks for visual recognition,'' \emph{IEEE transactions on
  pattern analysis and machine intelligence}, vol.~37, no.~9, pp. 1904--1916,
  2015.

\bibitem{liu2018path}
S.~Liu, L.~Qi, H.~Qin, J.~Shi, and J.~Jia, ``Path aggregation network for
  instance segmentation,'' in \emph{Proceedings of the IEEE conference on
  computer vision and pattern recognition}, 2018, pp. 8759--8768.

\bibitem{lin2014microsoft}
T.-Y. Lin, M.~Maire, S.~Belongie, J.~Hays, P.~Perona, D.~Ramanan,
  P.~Doll{\'a}r, and C.~L. Zitnick, ``Microsoft {COCO}: Common objects in
  context,'' in \emph{Proceedings of the European Conference on Computer Vision
  (ECCV)}.\hskip 1em plus 0.5em minus 0.4em\relax Springer, 2014.

\bibitem{arthur2006k}
D.~Arthur and S.~Vassilvitskii, ``{K-means}++: The advantages of careful
  seeding,'' Stanford, Tech. Rep., 2006.

\bibitem{luo2022target}
X.~Luo, Y.~Wu, and F.~Wang, ``Target detection method of {UAV} aerial imagery
  based on improved {YOLOv5},'' \emph{Remote Sensing}, vol.~14, no.~19, p.
  5063, 2022.

\bibitem{landau2011cluster}
S.~Landau, M.~Leese, D.~Stahl, and B.~S. Everitt, \emph{Cluster
  analysis}.\hskip 1em plus 0.5em minus 0.4em\relax John Wiley \& Sons, 2011.

\bibitem{ward1963hierarchical}
J.~H. Ward~Jr, ``Hierarchical grouping to optimize an objective function,''
  \emph{Journal of the American statistical association}, vol.~58, no. 301, pp.
  236--244, 1963.

\bibitem{zhou2017cad}
H.~Zhou, Z.~Li, C.~Ning, and J.~Tang, ``{CAD}: Scale-invariant framework for
  real-time object detection,'' in \emph{Proceedings of the IEEE International
  Conference on Computer Vision Workshops (ICCVW)}, 2017.

\bibitem{zhang2022focal}
Y.-F. Zhang, W.~Ren, Z.~Zhang, Z.~Jia, L.~Wang, and T.~Tan, ``Focal and
  efficient {IOU} loss for accurate bounding box regression,''
  \emph{Neurocomputing}, vol. 506, pp. 146--157, 2022.

\bibitem{nguyen2022towards}
H.~D. Nguyen, R.~Cai, H.~Zhao, A.~C. Kot, and B.~Wen, ``Towards more efficient
  security inspection via deep learning: A task-driven {X}-ray image cropping
  scheme,'' \emph{Micromachines}, vol.~13, no.~4, p. 565, 2022.

\end{thebibliography}

\end{document}